%% file: paper.tex
\documentclass[twoside]{article}
\usepackage{aistats2016}
\usepackage{subfigure}
\usepackage{longtable}
\usepackage{graphicx}
\usepackage{epstopdf}
\usepackage{algorithm}
\usepackage{algorithmic}
\usepackage{wrapfig}
\usepackage{hyperref}
\usepackage{color}
\usepackage{amsmath}
\usepackage{booktabs}
\usepackage{natbib}
\usepackage{appendix}

\include{defns}

\begin{document}

%
\runningtitle{UAVs using Bayesian Optimization}


\twocolumn[

\aistatstitle{UAVs using Bayesian Optimization to Locate WiFi Devices}

\aistatsauthor{ Mattia Carpin, Stefano Rosati, Mohammad Emtiyaz Khan, Bixio Rimoldi. }

\aistatsaddress{ \'Ecole Polytechnique F\'ed\'erale de Lausanne (EPFL), Switzerland. } ]

\begin{abstract}
We address the problem of localizing non-collaborative WiFi devices in a large region. Our main motive is to localize humans by localizing their WiFi devices, e.g. during search-and-rescue operations after a natural disaster. We use an active sensing approach that relies on Unmanned Aerial Vehicles (UAVs) to collect signal-strength measurements at informative locations. The problem is challenging since the measurement is received at arbitrary times and they are received only when the UAV is in close proximity to the device. For these reasons, it is extremely important to make prudent decision with very few measurements. We use the Bayesian optimization approach based on Gaussian process (GP) regression. This approach works well for our application since GPs give reliable predictions with very few measurements while Bayesian optimization makes a judicious trade-off between exploration and exploitation. In field experiments conducted over a region of 1000 $\times$ 1000 $m^2$, we show that our approach reduces the search area to less than 100 meters around the WiFi device within 5 minutes only. Overall, our approach localizes the device in less than 15 minutes with an error of less than 20 meters.
\end{abstract}

\section{Introduction}
\input{intro}

\section{Goals}
\input{goals.tex}

\section{Results}
\input{real_experiment.tex}
\begin{appendices}
\input{appendix.tex}
\end{appendices}

\small{
\bibliography{smavnet}
\bibliographystyle{plainnat}
}

\end{document}

%% file: defns.tex
\usepackage{subfigure}
\usepackage{epsfig}
\usepackage{amsmath,amsthm,amssymb}
\usepackage{url}
\usepackage{rotating}

\newcommand\cut[1]{}

\usepackage{algorithm}

\newcommand{\squishlist}{
   \begin{list}{$\bullet$}
    { \setlength{\itemsep}{0pt}      \setlength{\parsep}{3pt}
      \setlength{\topsep}{3pt}       \setlength{\partopsep}{0pt}
      \setlength{\leftmargin}{1.5em} \setlength{\labelwidth}{1em}
      \setlength{\labelsep}{0.5em} } }

\newcommand{\squishlisttwo}{
   \begin{list}{$\bullet$}
    { \setlength{\itemsep}{0pt}    \setlength{\parsep}{0pt}
      \setlength{\topsep}{0pt}     \setlength{\partopsep}{0pt}
      \setlength{\leftmargin}{2em} \setlength{\labelwidth}{1.5em}
      \setlength{\labelsep}{0.5em} } }

\newcommand{\squishend}{
    \end{list}  }
 
\newcommand{\real}{\mbox{$\mathbb{R}$}}

\newcommand{\sqr}[1]{\left[#1\right]}
\newcommand{\crl}[1]{\left\{#1\right\}}

\newcommand{\gauss}{\mbox{${\cal N}$}}

\newcommand{\myvec}[1]{\mbox{$\mathbf{#1}$}}
\newcommand{\myvecsym}[1]{\mbox{$\boldsymbol{#1}$}}

\newcommand{\vtheta}{\mbox{$\myvecsym{\theta}$}}

\newcommand{\vk}{\mbox{$\myvec{k}$}}

\newcommand{\vx}{\mbox{$\myvec{x}$}}

\newcommand{\vy}{\mbox{$\myvec{y}$}}

\newcommand{\vI}{\mbox{$\myvec{I}$}}

\newcommand{\vK}{\mbox{$\myvec{K}$}}

\newcommand{\vX}{\mbox{$\myvec{X}$}}

\newcommand{\calD}{\mbox{${\cal D}$}}

\newcommand{\data}{\calD}

\newcommand{\be}{\begin{equation}}
\newcommand{\ee}{\end{equation}}
\newcommand{\bea}{\begin{eqnarray}}
\newcommand{\eea}{\end{eqnarray}}
\newcommand{\beaa}{\begin{eqnarray*}}
\newcommand{\eeaa}{\end{eqnarray*}}

%% file: intro.tex
We consider localization of WiFi devices in a large region using an unmanned aerial vehicle (UAV). Due to the massive wide spread of commercial WiFi-enabled devices (smartphones, tablets, laptops, etc.), the location of WiFi devices can be used to localize people who own those devices.
Such localizations are important for many applications, e.g.  to deliver a mail package using UAVs for companies such as Amazon and Google, to provide internet connectivity using UAVs for companies like Facebook, and perhaps most importantly, in search and rescue operations to localize victims.


Localization of people could potentially be performed using bird's-eye view camera mounted on a UAV, as discussed in \cite{5509355}. 
%
%
However, this approach is limited to situations where the targets are clearly visible from the camera. Furthermore, it requires a great amount of image processing to separate the desired target from other objects.


%

Our approach is to localize WiFi devices measuring the received signal strength.
As specified by the 802.11x standards, a mobile-station (MS) periodically broadcasts a management frame, called Probe Request Frame (PRF), to actively scan the environment and discover access points. 
A WiFi device operating in monitor\footnote{More specifically, in RFMON (Radio Frequency MONitor) mode.} mode can receive and decode PRFs, extracting the MAC addresses of the source and measuring the corresponding signal strength index (RSSI).

The use of RSSI for localization purposes is problematic because there is no obvious connection between the measured signal strength and the distance between the transmitter and the receiver \cite{4604427}, \cite{5195701} and \cite{5267883}.

%

However, some authors have shown that the RSSI can actually be used in real situations to perform an estimation of the relative positions between the source and the receiving device \cite{4554465}. 
Methods using  Gaussian Processes (GP) are suited for modeling RSSI, since they can capture non-linearity and quantify uncertainty over a large region \cite{hahnel2006GauProSigStrLocEst}, \cite{ferris2007WiFUsiGauProLatVarMod}, \cite{5509574}. 
 
As shown in \cite{5195701}, by combining the RSSIs measured by several monitor nodes appropriately placed in a certain area, one can estimate and track the position of the MS transmitting the frames.
Such a method, however, is limited by the need of a large number of monitor nodes to cover large areas.

\begin{figure}[h!]
\centering
\includegraphics[width=0.8\columnwidth]{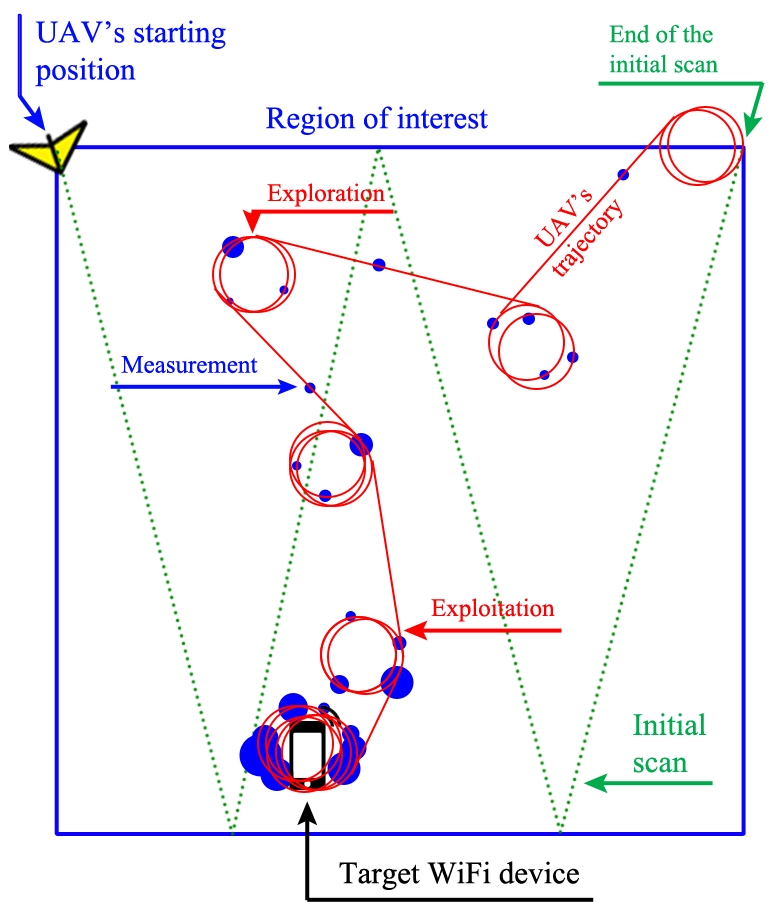}
\caption{Typical path followed by the UAV during the localization process. Blue circles show the data, and their size is proportional to the RSSI values.}
\label{fig:basic_setup}
\end{figure}

In this paper, we show that active sensing and exploitation using UAV results in a fast and accurate localization. Our approach is based on the fact that the expected value of an RSSI measurement increases as the distance decreases. The UAV therefore tries to get to a location where the expected value of the RSSI measurements is highest. 

Since collecting measurements is costly, both in terms of time and energy, we use the machine-learning method called Bayesian optimization. Such method allows the UAV to collect observations in informative locations to find the WiFi device quickly and with high accuracy.
More importantly, in our application, the environment is unknown and such ``black-box'' approach is appropriate since it avoids making assumptions about the environment. Our approach therefore is completely data-driven.   

Fig. \ref{fig:basic_setup} summarizes our approach for locating a single WiFi device. In the beginning, the UAV does a quick scan of the area to get an initial number of measurements. Based on such measurements, the UAV computes an estimate of the distribution of the RSSI  over the whole region. The UAV then decides to fly to a new location either to ``exploit", i.e., move to a location which is more likely to give higher RSSI measurements, or ``explore", i.e., it goes to previously unexplored locations.
This decision is taken using a \emph{Bayesian optimization} method. This process is repeated, i.e. as soon as a new measurement is made, the UAV moves to a new location either to exploit or to explore. The algorithm terminates when the confidence in our estimate is reasonable.

%% file: goals.tex
Our primary goal is to localize multiple WiFi devices in a given region. 
For simplicity, we will discuss localization of a single device. An extension to multiple devices is presented in Section \ref{sec:multiple_devices}. We will denote the true location of the device by $\vx^{true}$ in a region $\mathcal{X}$.

To localize the device, the UAV flies over the region $\mathcal{X}$ hoping to receive a measurement from the device. The measurements are received at random times. Let us suppose that by the time $t$, we have received $n_t$ measurements with time-stamps $t_1, t_2, \ldots, t_{n_t}$ such that $t_1<t_2<\ldots<t_{n_t}<t$. Denote by $\vx_{t_i}$ the location of the UAV when the $i$'th measurement was made and by $y_{t_i}$ the corresponding RSSI measurement. Note that the RSSI measurement is a random process $Y$  that depends on many factors such as the current position of UAV, the true location of the device and the time of measurement, but for notation simplicity we only show dependency w.r.t. the position of the UAV at time $t$, i.e. $\vx_t$.

Therefore, at any time $t$, we have a sequence of triplets: $\{t_i,\vx_{t_i},y_{t_i}\}$ for $i = 1,2,\ldots,n_t$ such that $t_i<t_{i+1}<t$. Denote the set of triplets at time $t$ by $\data_t$.

We will now discuss the characteristics of our data and the challenges associated with localization.

\section{Data Characteristics and Challenges}
The main characteristics of our data is that the measurement's strength is on average highest around the transmitting device. 

\begin{figure}[t!]
\centering
\includegraphics[width=3in]{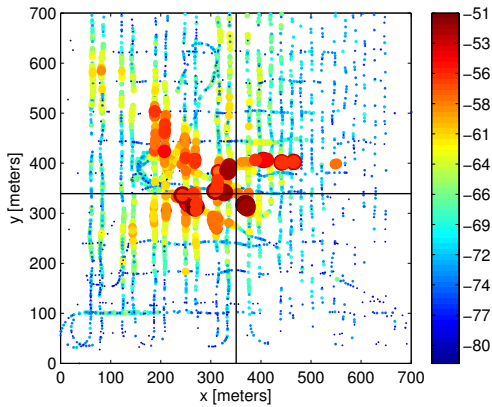}
\caption{RSSI measured during an exhaustive scan of the area. The device is located at the center (intersection of the horizontal and vertical lines). The size of a point is proportional to the signal strength.}
\label{fig:exhaustive_scan}
\end{figure}

Fig. \ref{fig:exhaustive_scan} shows the measurements gathered from a laptop (Sony Vaio Pro-13 ultrabook) during an exhaustive scan of the region that took more than 60 minutes. The device is located in the center of the figure, and the size and color of points are according to the signal strength. 

Unfortunately, localizing the peak is challenging due to the fact that the relation between the signal strength and the distance is stochastic.

\begin{figure}[t!]
\centering
\includegraphics[width=\columnwidth]{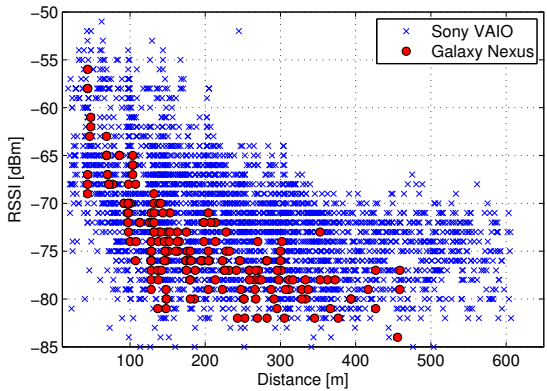}
\caption{RSSI as a function of the distance. Red points are measures gathered from a smartphone while blue points come from a laptop. }
\label{fig:RSSI_dist}
\end{figure}

This is because the measured signal strength is affected by noise, by the power of the transmitter, by the implementation of the receiver, and by the channel characteristic that include reflection, multipath scattering and shadowing. Therefore, at different distances, the distribution of the RSSI is different (thus making the measurements \emph{heteroscedastic}).
This is shown in Fig. \ref{fig:RSSI_dist} where we plot the signal strength as a function of distance for two devices. We clearly see that the two devices follow completely different profiles.

\begin{figure}[t!]
\centering
\includegraphics[width=\columnwidth]{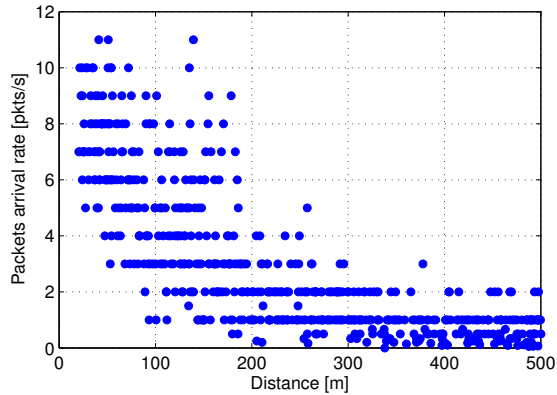}
\caption{Arrival measurements rate as a function of the distance.}
\label{fig:missingness}
\end{figure}

Secondly, since the device is non-collaborative and unaware of the localization, the measurements are received at random times.
More importantly, the chance of missing a measurement increases with the distance. This is shown in Fig. \ref{fig:missingness} where we plot the measurements arrival rate $1/(t_{i+1} - t_i)$ as a function of the distance from the device. 
Clearly, when we are close to the device, we get measurements more often compared to when we are far from it.
Therefore, in our problem the \emph{missing-at-random} \cite{book:rubin}, assumption is violated, making the problem challenging.

Finally, in our scenario, $\mathcal{X}$ could potentially be large and therefore it is crucial to develop an algorithm that is capable of collecting measurements where they really matter. We need to make good decisions with very few measurements, otherwise we may end up performing an exhaustive scan of $\mathcal{X}$ which will defeat the whole purpose.

As a result of these issues, naive methods such as hill-climbing do not work for our problem. Such methods needs a very good estimate of the gradient, and such estimate would require far more measurements that the ones that we have.

\section{Our Approach}
The UAV first performs a rough scan of the whole area looking for the an initial number of RSSI measurements. This initial phase, in our scenario, takes around 3 minutes. As soon as the initial phase is completed, it starts the process of localization in which it repeatedly performs the following tasks whenever a new measurement is received.
\begin{enumerate}
\item Using the data $\data_t$ at time $t$, the UAV computes the (predictive) distribution $p(y | \vx_*, \data_t)$. This is the probability density function of the measurements $y$ at the candidate locations $\vx_*\in\mathcal{X}$ given the past measurements $\data_t$. 
Gaussian process regression is used for this purpose.
\item Given the distribution, it decides 
whether to \emph{exploit} or to \emph{explore} based on the \emph{expected improvement} function. This step is based on the Bayesian-optimization framework \cite{bayes_opt}.
\item The UAV moves then to the new location and waits for a new measurement. When a new measurement it received, it updates the estimate of the true location and decides whether to terminate the search or to repeat the three steps.
\end{enumerate}
Algorithm 1 contains the details.

\begin{algorithm}[t]
\label{alg:localization_single}
\caption{Localization of a single WiFi device}
\begin{algorithmic}[1]
\STATE Do an initial scan as described in Section \ref{sec:issues} to get an initial number of measurements.
\REPEAT
  \STATE Randomly sample $n_*$ candidate locations $\vx_*$ and compute distribution of RSSI $y$ at those location using GP (see Eq. \ref{eq:GPmean} and \ref{eq:GPvariance}).
  \STATE Go to the location that maximizes EI (Eq. \ref{eq:EI_gauss}).
  \STATE Update the estimate of device location, when a new measurement is received. 
  \STATE Terminate when estimate don't change much, otherwise repeat.
\UNTIL{terminate}
\end{algorithmic}
\end{algorithm}

\subsection{Distribution computation using Gaussian Process Regression}
Gaussian process (GP) regression is one of the most popular choice within the Bayesian optimization framework. It is also particularly suited for our application for many reasons. 
%
%
Firstly, GP regression works well in small-sample setting which is desirable in our setting since in the beginning of the search we do not have a lot of data. Secondly, GP regression not only gives us predictions but also uncertainty associated with them, which helps us choose between exploration and exploitation (see more details in the next section). Finally, hyperparameters of GP regression can be set easily using maximum likelihood estimation, allowing for an easy implementation (see end of this section).

In this paper, we use a regression model that gives us a Gaussian  distribution at each candidate location $\vx_*$. 
In this paper, we assume that the measurements are sampled from a jointly Gaussian distribution.
We make this choice since it leads to a computationally simple algorithm which can be implemented on the on-board computer in the UAV. Our model therefore takes the following form: 
\begin{align}
y_t &= f_t + \epsilon_t, \quad \textrm{ for } t = t_1,t_2,\ldots \label{eq:meas_model}
\end{align}
where $\epsilon_t$ are i.i.d. Gaussian noise $\epsilon_t \sim \gauss(0,\sigma^2)$ with variance $\sigma^2$, while $f_t$ is drawn from a GP $f_t \sim \mathcal{GP}(m(\vx), k(\vx,\vx'))$ with mean function $m$ and covariance function $k$. The relationship between $y$ and $\vx$ is captured with a specific mean and covariance function. In this paper, we use the zero mean function and a squared-exponential covariance function defined below:
\begin{align}
k(\vx,\vx^{'}) &= \sigma_f^2 \exp\sqr{-\frac{1}{l^2}(\vx-\vx^{'})^T (\vx-\vx^{'})}
\end{align}
where $\sigma_f,l\in\real$ are kernel hyperparameters that control the  spatial correlation. We denote the set of hyperparameters by $\vtheta = \{\sigma, \sigma_f, l\}$.

The distribution $p(y|\vx_*,\data_t, \vtheta)$ at a candidate location $\vx_*$ is a Gaussian distribution. Denote the vector of measurements received until time $t$ by $\vy_t = [y_{t_1}, y_{t_2}, \ldots, y_{t_{n_t}}]^T$.
%
It follows from the property of GPs that any finite number of samples drawn from GP are jointly Gaussian, giving the following expression for the distribution of $\vy_t$ and any $y$: 
\begin{align}
\begin{bmatrix}
\vy_t \\ 
y 
\end{bmatrix}
\sim
\gauss
\Bigg(
\mathbf{0},\begin{bmatrix}
\vK_t+\sigma^2\vI & \vk_* \\ 
\vk_*^T & k_{**} + \sigma^2
\end{bmatrix}\Bigg)
\end{align}
where $\vK_t$ is a matrix with $(i,j)$'th entry as $k(\vx_{t_i},\vx_{t_j})$, $\vk_*$ is a vector with $i$'th entry as $k(\vx_{t_i},\vx_*)$ and $k_{**} = k(\vx_*,\vx_*)$.

We can write the expression for the distribution of $y$ given $\vy_t$ (see page 16 of \cite{book:rasmussen}):
\begin{align}
p(y|\vx_*,\data_t, \vtheta)&:= \gauss(\mu_{*|t}, \sigma^2_{*|t}) \\
\textrm{ where } \mu_{*|t} &:= \vk_*^T (\vK_t + \sigma^2\vI)^{-1} \vy_t \label{eq:GPmean}\\
\sigma^2_{*|t} &:= k_{**} - \vk_*^T (\vK_t + \sigma^2\vI)^{-1} \vk_* \label{eq:GPvariance}
\end{align}
The computational complexity of these operations is $O(n_* n_t^3)$ where $n_*$ is the number of candidate locations. We limit the number of candidate locations to around $n_*=350$. The cubic cost can be reduced by using one rank updates. Nevertheless, this computation is acceptable in our application since we do not have to measure more than 300 measurements for the whole search operation.

The distribution  $p(y|\vx_*,\data_t, \vtheta)$ depends on specification of $\vtheta$. 
We set $\vtheta$ offline using the data from a scan of the region (such as Fig. \ref{fig:exhaustive_scan}). Specifically, given a measurement vector $\vy$ taken at locations $\vX$, we maximize $\log p(\vy|\vX,\vtheta)$. The expression for the log-likelihood is also available in closed-form. Please see \cite{book:rasmussen} for details.

\subsection{Deciding UAV's Next Location using Bayesian Optimization}
We use the Bayesian optimization framework to decide the next location to visit. The decision in made based on the distribution $\gauss(\mu_{*|t}, \sigma^2_{*|t})$. Intuitively, the decision trades-off the two competing goals:
\begin{enumerate}
\item The UAV should visit regions that are similar to regions where high RSSI measurements were observed. This is the \emph{exploitation} step.
\item The UAV should also visit unexplored regions that could also potentially give rise to high RSSI measurements. This is the \emph{exploration} step.
\end{enumerate}
Bayesian optimization takes these two goals into account by using an \emph{acquisition function} \cite{bayes_opt}. In this paper, we use the expected-improvement (EI) acquisition function which measures the expected improvement over the location where the last highest RSSI measurement was observed. Let $\mu^+$ be the greatest mean value among those computed $\forall \vx \in \mathcal{D}_t$.  Then the EI function is defined as follows:
\begin{align}
\textrm{EI}(\vx_*) &:= \mathbb{E}_{p(y|\mathbf{x}_*,\mathcal{D}_t)} \sqr{ \max \crl{ 0, \mu_{*|t} - \mu^+ }}  
\end{align}
For Gaussian predictive distribution, EI has a closed-form expression:
\begin{align}
\textrm{EI}(\vx_*) &= (\mu_{*|t} - \mu^+) \Phi(u)+\sigma_{*|t} \phi(u) \label{eq:EI_gauss}
\end{align}
where $u :=(\mu_{*|t} - \mu^+)/ \sigma_{*|t}$ and we assume that $\sigma_{*|t}>0$, $\Phi$ and $\phi$ are cumulative distribution function and probability density function of the standard Gaussian respectively. When $\sigma_{*|t}$ is 0, we set EI to be zero as well. This equation shows that EI will be large under two circumstances: either the test location has a high mean value $\mu_{*|t}$ or it has a high $\sigma_{*|t}$. Clearly, the two scenarios correspond to exploitation and exploration respectively.

Using the distribution, we can compute an estimate of $x^{true}$. We use for such estimate the $\vx_*$ for which $\mu_{*|t}$ is maximum.

\subsection{Practical Issues}  \label{sec:issues}

{\bf Initialization using a quick scan:} In the initialization phase, the UAV flies with a predetermined trajectory. Starting from the upper-left corner of the search area, it reaches the opposite corner bouncing from one edge to the other $n_{init}$ times. 
The goal of this phase is to collect an initial number of measurement in order to start the Bayesian optimization algorithm. 

{\bf Stopping rule:} The algorithm stops when we don't make any improvement in our estimate. In particular, we stop when a few successive predictions fall inside the same neighborhood of some predetermined radius $\delta$.


{\bf Minor details about navigation:} There are two minor details for task 2 specific to our set-up. Firstly, the UAV we used in our experiment does not exactly go to the new specified location, rather it flies in a circle \emph{around} this location with a pre-defined radius. However, the location $\vx_{t}$ where measurements $y_t$ is received is the actual position of the UAV at time $t$. Secondly, having made a decision to move to a new location, the UAV does not make any new decisions until it reaches to the new location. New measurements received during the flight are added to the dataset and are used for future decisions.

\section{Details of UAV}

We can distinguish two main types of small UAVs: rotary-blade (i.e. multi-copters) and fixed-wing.
Multi-copters fly using the lift force generated by one or more blades (small UAVs have typically four blades, thus they are called quadricopters)
They can take off and land vertically, and hover in the air.
However, they are slow, particularly sensitive to the wind, and therefore unsuited to cover long distances. 
In order to cover a large search areas, we adopted fixed-wing UAVs, because they fly at higher speeds and with a higher energy efficiency compared to quadricopters.

\begin{figure}[t!]
\centering
\includegraphics[width=\columnwidth]{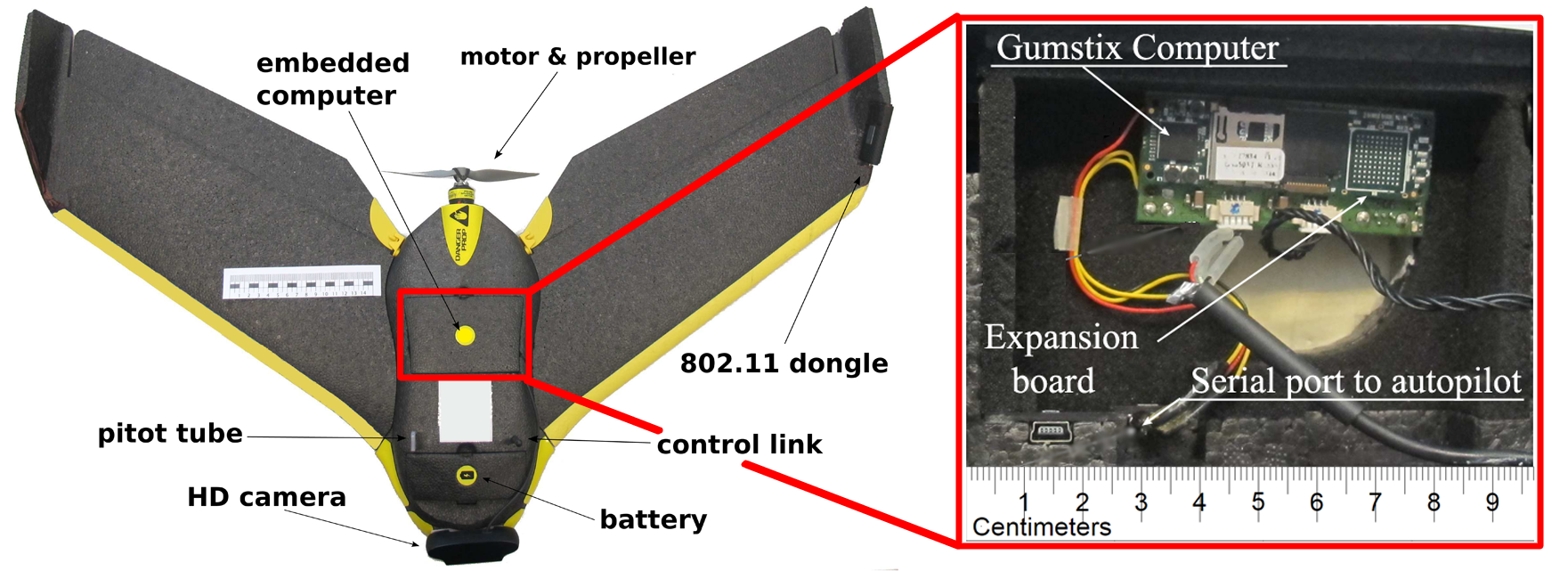}
\caption{UAV platform used in the experiments. Left figure shows the SenseFly \emph{eBee} with HD camera, Linksys AE3000 USB dongle, and Gumstix computer-on-module. Right figure shows the computer and expansion board located in the rear compartment of the UAV.}
\label{fig:ebee}
\end{figure}

For the experiments we used a fixed-wing UAV, named \emph{eBee}, developed by SenseFly \cite{bib:sensefly}. 
 The vehicle's body is made of expanded polypropylene (EPP), and it has a single rear-mounted propeller powered by an electric motor. The \emph{eBee} platform is illustrated in Fig. \ref{fig:ebee}.
It has an integrated autopilot capable of flying with winds up to 12 m/s, at a cruising speed of about 57 km/h, with an autonomy of 45 minutes. 
%

We installed an ARM-based computer-on-module named Gumstix Overo Tide \cite{bib:overoTide} on the rear compartment of the plane, where the localization algorithm runs.
The computer runs a customized  Linux distribution (kernel version 3.5.0) and it is connected with two 802.11 radio interfaces via USB, and with the autopilot via serial port. 
The first 802.11 interface is set in monitor mode to sniff PRFs. The second one is used to connect the UAV computer to our ground station.
The Gumstix computer interacts with the onboard autopilot to fly the UAV over specific locations decided from the localization algorithm.

%% file: real_experiment.tex
\subsection{Hyperparameter Estimation for GP}
As discussed earlier, the chosen hyperparameters are those that maximize the log likelihood $\log p(\vy|\vX,\vtheta)$. They are $l=0.2$, $\sigma=4.47$ and $\sigma_f=3.32$. 

\begin{figure}[t]
\centering
\includegraphics[width=\columnwidth]{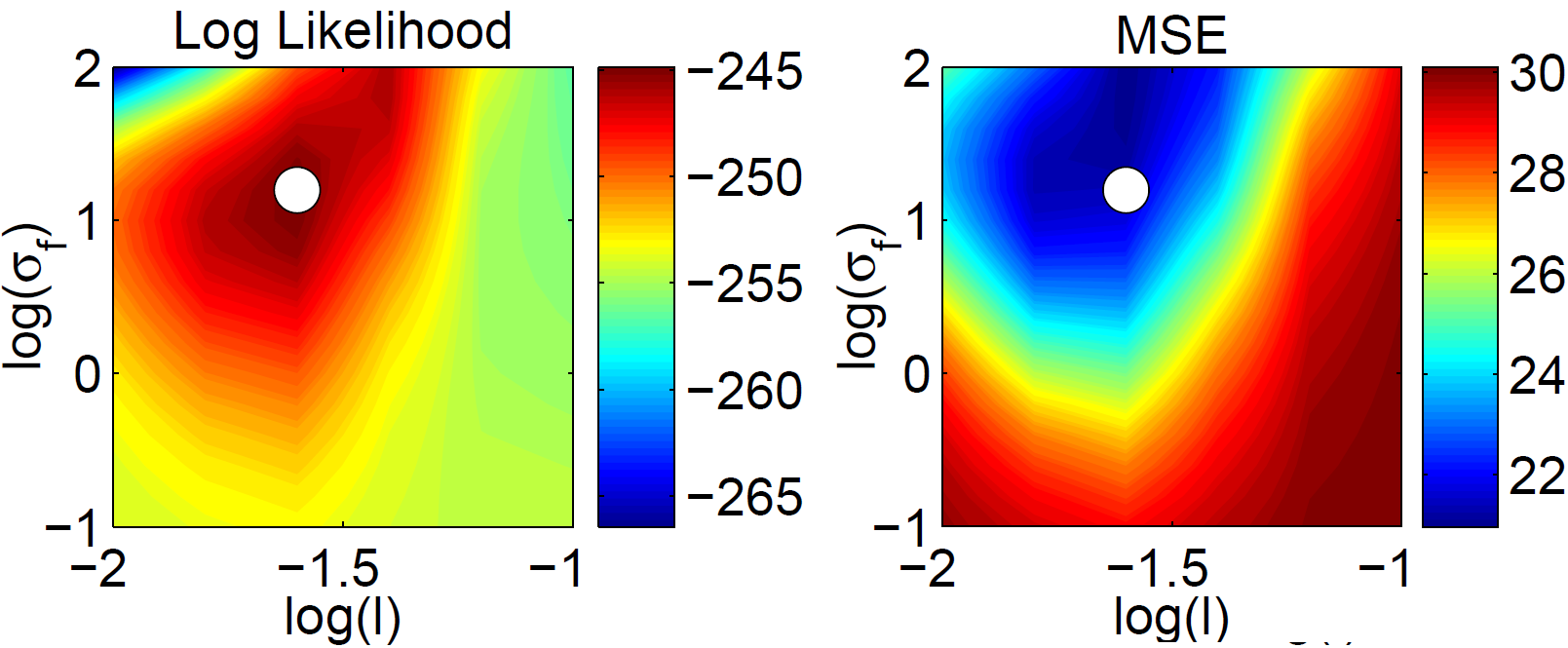}
\caption{Marginal log-likelihood and MSE for different values of the hyperparameters $(l,\sigma_f)$. The white circle in both plots denotes the maximum value of the log likelihood.}
\label{fig:traintest}
\end{figure}

To verify that we are not overfitting the data, we have tested the chosen values against new measurements. 
Figure \ref{fig:traintest} (right) shows the MSE between new samples and the mean value that follows from the chosen hyperparameters. The left side of Figure \ref{fig:traintest} shows the log likelihood.
We can clearly see that the hyperparameters that maximize the log likelihood also give a very low value of MSE, justifying the choice.

\subsection{Real-Data Based Simulations}
This section reports simulation results, based on real measurements. Our results suggest that the proposed method is able to localize the device with an average error of 50 meters within 5 minutes. Specifically, we simulate the following:
\begin{enumerate}
\item The trajectory followed by the UAV during the localization process.
\item The RSSI measurements.
\item The random times at which the signals are emitted. 
\end{enumerate}
The details of the last two are explained in the next two paragraphs.

The RSSI measurements are sampled from an empirical distribution obtained from the real data show in in Figure \ref{fig:RSSI_dist}. Specifically, the range between $[0,600]$ meters has been divided in 20 bins. For each bin, we have derived the corresponding empirical distribution. This distribution is used to draw a RSSI sample for any distance that falls into the bin. 


The times at which PRFs are transmitted are selected independently. The intertransmittion interval is a random variable uniformly distributed in the interval $[\Delta T-\varepsilon_{max},\Delta T+\varepsilon_{max}]$.

\begin{figure}[t]
\centering
\includegraphics[width=\columnwidth]{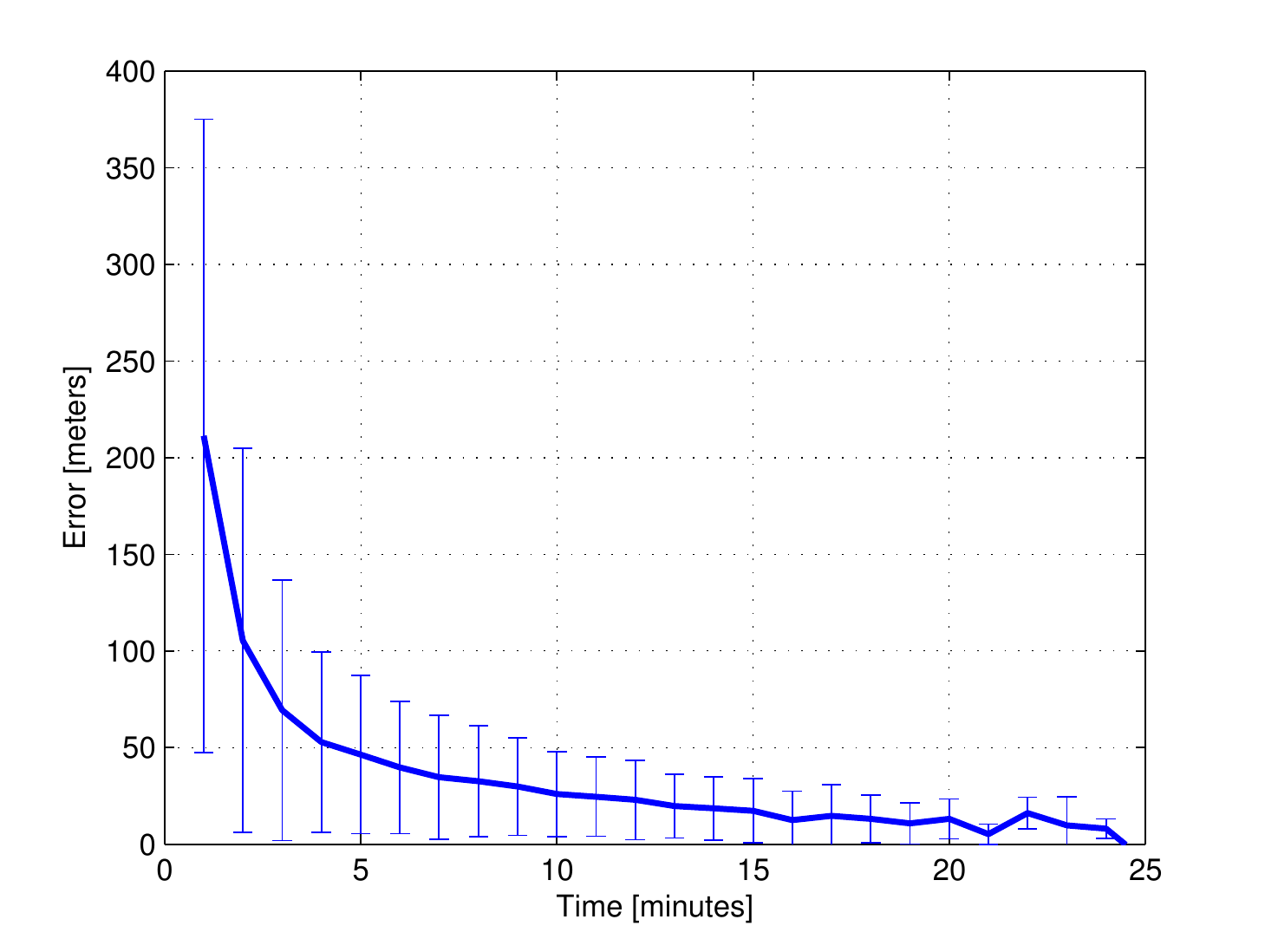}
\caption{Mean localization error plus and minus standard deviation as a function of the time.}
\label{fig:errorbar}
\end{figure}

Figure \ref{fig:errorbar} shows the average localization error and the standard deviation as a function of the time, obtained from 1000 independent runs of the algorithm.

These results suggest that within 5 minutes the UAV is able to reach within 100 meters of the device. This is a huge improvement over the time taken to do a full scan of the region, e.g. the scan of Figure \ref{fig:exhaustive_scan} took around an hour. 
In 5 minutes one would typically get only 10 to 30 RSSI measurements. According to the figure the average error at this time is below 50 meters.

\begin{figure}[t]
\centering
\includegraphics[width=\columnwidth]{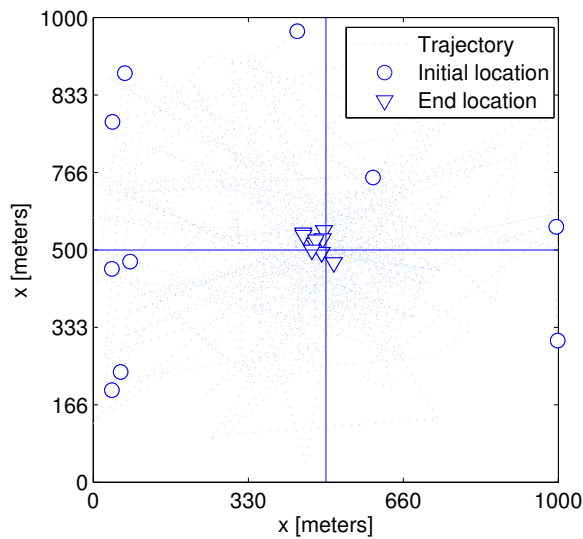}
\caption{Trajectories followed by the UAV during 10 runs of the localization process. The source is at the intersection of the horizontal and vertical line.}
\label{fig:trajectories}
\end{figure}

Figure \ref{fig:trajectories} shows the UAV's trajectory during 10 independent runs. The initial locations are shown by a circle while the end locations are shown with a triangle. In every run the initial location was different, yet all final locations where within 30 meters of the transmitter. Given that in the final phase is an exploitation phase, the location of the UAV corresponds to the estimated WiFi device position.

\subsection{Real-World Experiments}

\begin{figure}[t]
\centering
\includegraphics[width=\columnwidth]{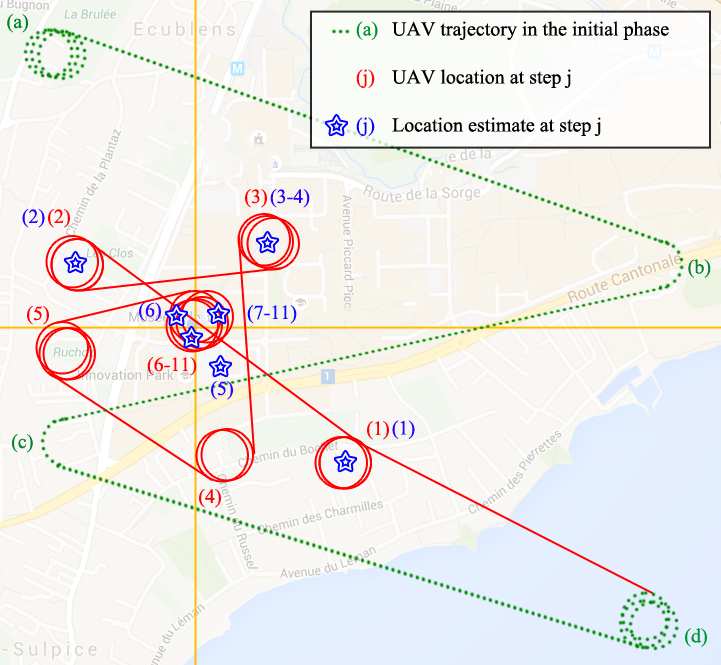}
\caption{Trajectory followed by the UAV during the localization process. The source is at the intersection of the horizontal and vertical line.}
\label{fig:map}
\end{figure} 

\begin{figure}[h]
\centering
\includegraphics[width=0.9\columnwidth]{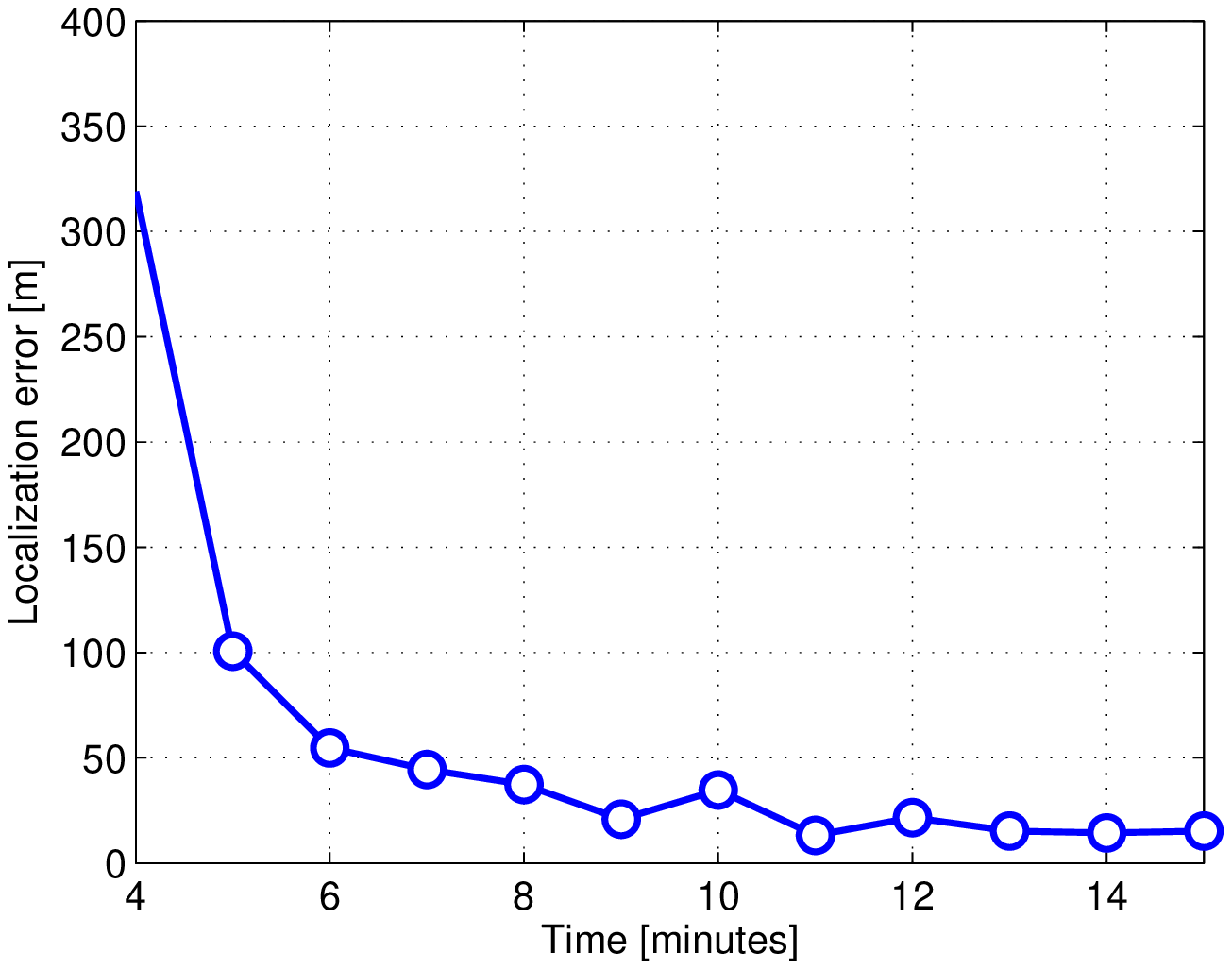}
\caption{Localization error vs time for the real-world experiment.}
\label{fig:real_error}
\end{figure} 

In this section, we present results of a field experiment.
The WiFi device was a smartphone Samsung Galaxy Nexus i9250, positioned in a square of 1000 $\times$ 1000 $m^2$. 

The UAV followed the trajectory shown in Figure \ref{fig:map}.  Figure \ref{fig:real_error} shows the localization error vs time. We see that within 6 minutes the was error is reduced to 50 meters, validating the simulation results.
The whole flight took approximately 15 minutes, after which the localization error was 18 meters.

\begin{figure}[h]
\centering
\begin{subfigure}[$n_t=7$, $t=$ 5:05]{
\includegraphics[width=\columnwidth]{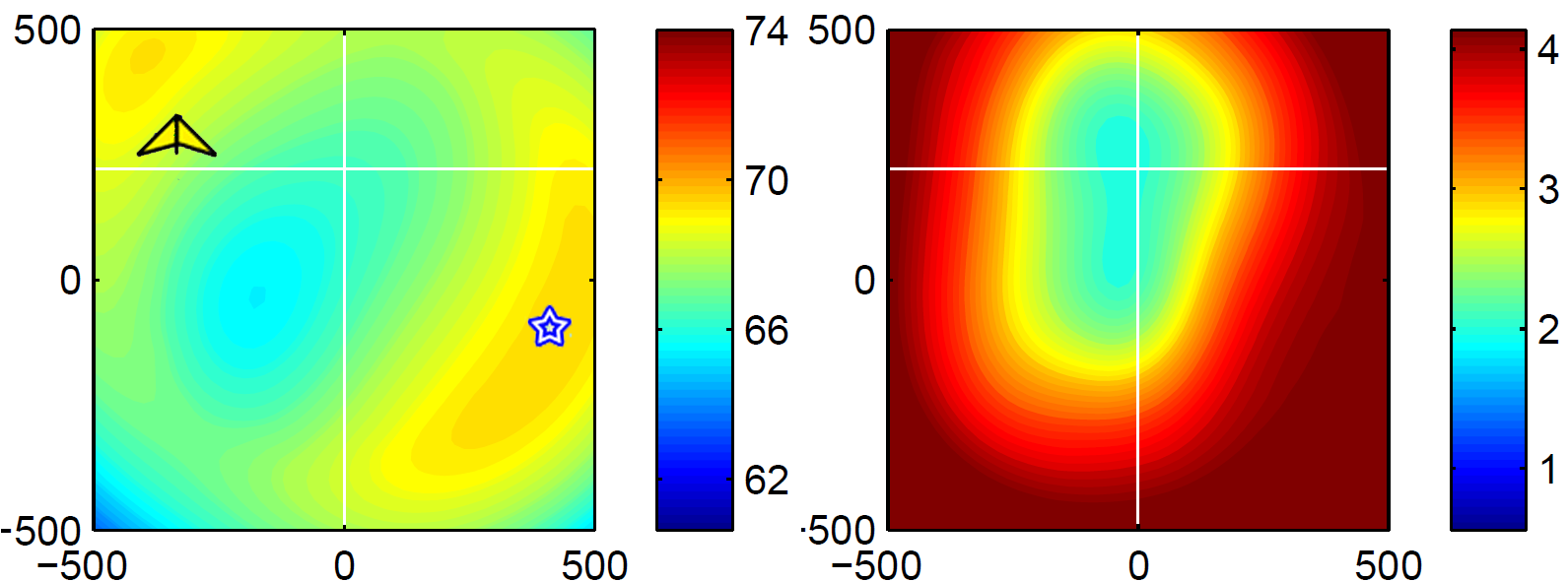}}
\end{subfigure} 
\begin{subfigure}[$n_t=16$, $t=$ 8:52]{
\includegraphics[width=\columnwidth]{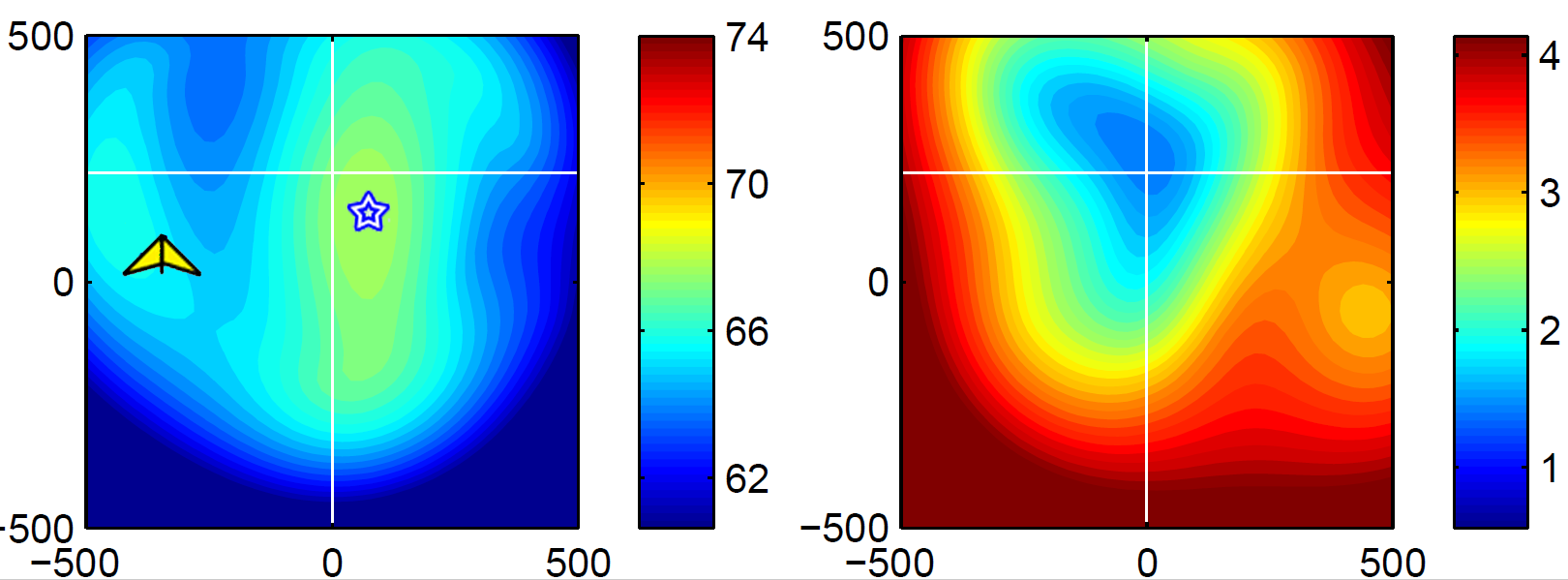}}
\end{subfigure} 
\begin{subfigure}[$n_t=47$, $t=$ 14:48]{
\includegraphics[width=\columnwidth]{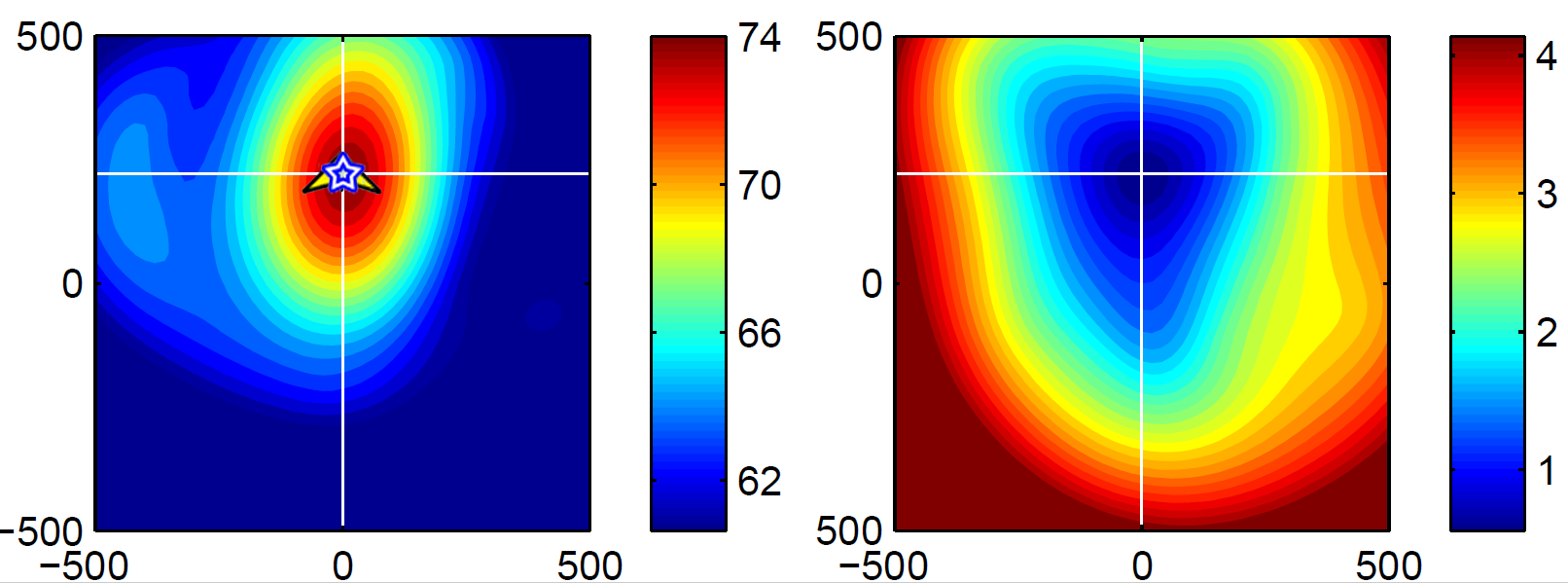}}
\end{subfigure} 
\caption{Posterior mean (left columns) and standard deviation (right column) at different times of the localization process. The location of the target is indicated with the white cross.}
\label{fig:post_mean_var}
\end{figure}

Figure \ref{fig:post_mean_var} shows the evolution of means and variances of $y$, conditioned on the measurements. From the same Fiugre, we can also see whether the UAV is in exploration or exploitation phase.

After 5 minutes, the UAV had collected only 7 measurements.
The predicted mean at that time was almost flat with a high variance almost everywhere (see Figure \ref{fig:post_mean_var}(a)). The star indicates the estimated device position, while the other marker shows where  the UAV is going to collect the next measurement.
As we can see, the UAV decides to do an exploration of the region moving away from the peak (shown with star).

Figures \ref{fig:post_mean_var}(b) and \ref{fig:post_mean_var}(c) show predictions after roughly 9 and 15 minutes respectively. We observe that as time progresses, the mean concentrates around the true location of the device and the variance at the same location reduces. We also see that the UAV keeps exploring after roughly 9 mins and does an exploitation step after 15 minutes.
The algorithm terminates at $t=15$ minutes, with a localization error of 18 meters.

\section{Conclusions}
In conclusion, both the field experiments and simulation results confirmed the effectiveness of the Bayesian optimization approach combined with GP regression to locate a non collaborative WiFi device that emits PRF sparsely.
Specifically, in the field experiment, after 15 minutes the estimate was within 18 meters of the actual device.

%% file: appendix.tex
\section{Appendix}
\subsection{Extension to Multiple Devices}
\label{sec:multiple_devices}
The sequential extension of the algorithm described in the previous section is straightforward.
After having located the first device, the algorithm repeats the Bayesian driven search for each other target, while the initialization phase is performed only once at the
beginning. During the localization of the first target, the UAV can potentially collect packets coming from any other target. Therefore the algorithm has an advantage when the estimation of the second target's location starts, because the dataset is already rich. 

To improve the localization efficiency, we designed a specific variant of the algorithm to locate multiple devices. The idea is to fly the UAV in specific locations to collect measurements that can be useful for localizing all the $M$ devices.
Let us define $\mathcal{F}$ as the set containing the indexes of the devices not yet located. After the initialization phase, $\mathcal{F}=\lbrace 1,2,\dots,M \rbrace$.
As a first step, we derive independently the distribution $\gauss_j(\mu_{{*,j}|t}, \sigma^2_{{*,j}|t})$, $\forall j \in \mathcal{F}$. Then we compute the aggregated distribution $\tilde{\gauss}(\tilde{\mu}_{*,t},\tilde{\sigma}_{*,t})$ according to

\begin{equation}
\left\{
\begin{array}{l}
\tilde{\mu}_{*,t}=\smash{\displaystyle\max_{j \in \mathcal{F}}} ( \mu_{{*,j}|t})\\\\
\tilde{\sigma}_{*,t}=\smash{\displaystyle\max_{j \in \mathcal{F}}} ( \sigma_{{*,j}|t})\\
\end{array}
\right.
\end{equation}
The \textit{max} operator has been used to aggregate all the distribution in order to preserve the peaks, that are crucial for a precise estimation. The acquisition function makes use of $\tilde{\gauss}(\tilde{\mu}_{*,t},\tilde{\sigma}_{*,t})$ to decide where the UAV moves next. Once device $i$ has been located, then $\mathcal{F}=\mathcal{F} \backslash \lbrace i \rbrace$.

%% file: paper.bbl
\begin{thebibliography}{13}
\providecommand{\natexlab}[1]{#1}
\providecommand{\url}[1]{\texttt{#1}}
\expandafter\ifx\csname urlstyle\endcsname\relax
  \providecommand{\doi}[1]{doi: #1}\else
  \providecommand{\doi}{doi: \begingroup \urlstyle{rm}\Url}\fi

\bibitem[Benkic et~al.(2008)Benkic, Malajner, Planinsic, and Cucej]{4604427}
K.~Benkic, M.~Malajner, P.~Planinsic, and Z.~Cucej.
\newblock {Using RSSI value for distance estimation in wireless sensor networks
  based on ZigBee}.
\newblock In \emph{{Systems, Signals and Image Processing, 2008. IWSSIP 2008.
  15th International Conference on}}, pages 303--306, June 2008.
\newblock \doi{10.1109/IWSSIP.2008.4604427}.

\bibitem[Brochu et~al.(2010)Brochu, Cora, and {de Freitas}]{bayes_opt}
Eric Brochu, Vlad~M Cora, and Nando {de Freitas}.
\newblock {A Tutorial on Bayesian Optimization of Expensive Cost Functions,
  with Application to Active User Modeling and Hierarchical Reinforcement
  Learning}.
\newblock eprint arXiv:1012.2599, arXiv.org, December 2010.

\bibitem[Ferris et~al.(2007)Ferris, Fox, and
  Lawrence]{ferris2007WiFUsiGauProLatVarMod}
Brian Ferris, Dieter Fox, and Neil~D Lawrence.
\newblock {WiFi-SLAM Using Gaussian Process Latent Variable Models.}
\newblock In \emph{{IJCAI}}, volume~7, pages 2480--2485, 2007.

\bibitem[Fink and Kumar(2010)]{5509574}
J.~Fink and V.~Kumar.
\newblock {Online methods for radio signal mapping with mobile robots}.
\newblock In \emph{{Robotics and Automation (ICRA), 2010 IEEE International
  Conference on}}, pages 1940--1945, May 2010.
\newblock \doi{10.1109/ROBOT.2010.5509574}.

\bibitem[Gumstix()]{bib:overoTide}
Gumstix.
\newblock {Gumstix Web Page}.
\newblock https://www.gumstix.com.

\bibitem[H\"{a}hnel and Fox(2006)]{hahnel2006GauProSigStrLocEst}
Brian Ferris~Dirk H\"{a}hnel and Dieter Fox.
\newblock {Gaussian Processes for Signal Strength-Based Location Estimation}.
\newblock In \emph{{Proceeding of Robotics: Science and Systems}}, 2006.

\bibitem[Jianwu and Lu(2009)]{5267883}
Zhang Jianwu and Zhang Lu.
\newblock {Research on distance measurement based on RSSI of ZigBee}.
\newblock In \emph{{Computing, Communication, Control, and Management, 2009.
  CCCM 2009. ISECS International Colloquium on}}, volume~3, pages 210--212, Aug
  2009.
\newblock \doi{10.1109/CCCM.2009.5267883}.

\bibitem[Little and Rubin(1986)]{book:rubin}
Roderick J~A Little and Donald~B Rubin.
\newblock \emph{Statistical Analysis with Missing Data}.
\newblock John Wiley \& Sons, Inc., New York, NY, USA, 1986.
\newblock ISBN 0-471-80254-9.

\bibitem[Rasmussen and Williams(2006)]{book:rasmussen}
Carl~Edward Rasmussen and Christopher K.~I. Williams.
\newblock \emph{{Gaussian Processes for Machine Learning}}.
\newblock The MIT Press, 2006.
\newblock ISBN 026218253X.

\bibitem[Saxena et~al.(2008)Saxena, Gupta, and Jain]{4554465}
M.~Saxena, P.~Gupta, and B.N. Jain.
\newblock {Experimental analysis of RSSI-based location estimation in wireless
  sensor networks}.
\newblock In \emph{{Communication Systems Software and Middleware and
  Workshops, 2008. COMSWARE 2008. 3rd International Conference on}}, pages
  503--510, Jan 2008.
\newblock \doi{10.1109/COMSWA.2008.4554465}.

\bibitem[SenseFly()]{bib:sensefly}
SenseFly.
\newblock {senseFly eBee UAV webpage}.
\newblock http://www.sensefly.com/drones/ebee.html.
\newblock Accessed: 2015-10-09.

\bibitem[Symington et~al.(2010)Symington, Waharte, Julier, and
  Trigoni]{5509355}
A.~Symington, S.~Waharte, S.~Julier, and N.~Trigoni.
\newblock {Probabilistic target detection by camera-equipped UAVs}.
\newblock In \emph{{Robotics and Automation (ICRA), 2010 IEEE International
  Conference on}}, pages 4076--4081, May 2010.
\newblock \doi{10.1109/ROBOT.2010.5509355}.

\bibitem[Zarimpas et~al.(2006)Zarimpas, Honary, and Darnell]{5195701}
Vasileios Zarimpas, Bahram Honary, and Mike Darnell.
\newblock {Indoor 802.1x based location determination and real-time tracking}.
\newblock In \emph{{Wireless, Mobile and Multimedia Networks, 2006 IET
  International Conference on}}, pages 1--4, Nov 2006.

\end{thebibliography}
